\documentclass[letterpaper]{article}
\usepackage{aaai}
\usepackage{times}
\usepackage{helvet}
\usepackage{courier}
\usepackage{hyperref}
\usepackage{verbatim}
\usepackage{outlines}
\usepackage{graphicx}
\usepackage{color}
\usepackage{xcolor}
\usepackage{paralist}
\copyrightyear{2020}
\usepackage{float}
\usepackage{tikz-cd}
\tikzcdset{diagrams={nodes={inner sep=2pt}}}
\usepackage{multirow}
\usepackage{romannum}
\usepackage{amsmath,amssymb}
\usepackage[ruled]{algorithm2e} %linesnumbered, 
\SetKwInOut{Input}{Input}
\SetKwInOut{Output}{Output\,}
\SetKwInOut{Initialize}{Initialize}
\SetKwInOut{Data}{Data}
\usepackage{diagbox, eqparbox, hhline}
\usepackage{caption}
\usepackage{subcaption}
\usepackage{cleveref}

\newcommand{\textblack}[1]{\textcolor{black}{{#1}}}
\newcommand{\indep}{\rotatebox[origin=c]{90}{$\models$}}
\begin{document}
 \setcounter{secnumdepth}{2}

\def\Blue{\color{blue}}
\def\Purple{\color{purple}}

\def\A{{\bf A}}
\def\a{{\bf a}}
\def\B{{\bf B}}
\def\b{{\bf b}}
\def\C{{\bf C}}
\def\c{{\bf c}}
\def\D{{\bf D}}
\def\d{{\bf d}}
\def\E{{\bf E}}
\def\e{{\bf e}}
\def\f{{\bf f}}
\def\F{{\bf F}}
\def\K{{\bf K}}
\def\k{{\bf k}}
\def\L{{\bf L}}
\def\h{{\bf h}}
\def\G{{\bf G}}
\def\g{{\bf g}}
\def\I{{\bf I}}
\def\R{{\bf R}}
\def\X{{\bf X}}
\def\Y{{\bf Y}}
\def\OO{{\bf O}}
\def\oo{{\bf o}}
\def\P{{\bf P}}
\def\Q{{\bf Q}}
\def\r{{\bf r}}
\def\s{{\bf s}}
\def\S{{\bf S}}
\def\t{{\bf t}}
\def\T{{\bf T}}
\def\x{{\bf x}}
\def\y{{\bf y}}
\def\z{{\bf z}}
\def\Z{{\bf Z}}
\def\M{{\bf M}}
\def\m{{\bf m}}
\def\n{{\bf n}}
\def\U{{\bf U}}
\def\u{{\bf u}}
\def\V{{\bf V}}
\def\v{{\bf v}}
\def\W{{\bf W}}
\def\w{{\bf w}}
\def\0{{\bf 0}}
\def\1{{\bf 1}}

\def\AM{{\mathcal A}}
\def\EM{{\mathcal E}}
\def\FM{{\mathcal F}}
\def\TM{{\mathcal T}}
\def\UM{{\mathcal U}}
\def\XM{{\mathcal X}}
\def\YM{{\mathcal Y}}
\def\NM{{\mathcal N}}
\def\OM{{\mathcal O}}
\def\IM{{\mathcal I}}
\def\GM{{\mathcal G}}
\def\PM{{\mathcal P}}
\def\LM{{\mathcal L}}
\def\MM{{\mathcal M}}
\def\DM{{\mathcal D}}
\def\SM{{\mathcal S}}
\def\RB{{\mathbb R}}
\def\EB{{\mathbb E}}

\def\tx{\tilde{\bf x}}
\def\ty{\tilde{\bf y}}
\def\tz{\tilde{\bf z}}
\def\hd{\hat{d}}
\def\HD{\hat{\bf D}}
\def\hx{\hat{\bf x}}
\def\hR{\hat{R}}

\def\Ome{\mbox{\boldmath$\omega$\unboldmath}}
\def\bet{\mbox{\boldmath$\beta$\unboldmath}}
\def\et{\mbox{\boldmath$\eta$\unboldmath}}
\def\ep{\mbox{\boldmath$\epsilon$\unboldmath}}
\def\ph{\mbox{\boldmath$\phi$\unboldmath}}
\def\Pii{\mbox{\boldmath$\Pi$\unboldmath}}
\def\pii{\mbox{\boldmath$\pi$\unboldmath}}
\def\Ph{\mbox{\boldmath$\Phi$\unboldmath}}
\def\Ps{\mbox{\boldmath$\Psi$\unboldmath}}
\def\pss{\mbox{\boldmath$\psi$\unboldmath}}
\def\tha{\mbox{\boldmath$\theta$\unboldmath}}
\def\Tha{\mbox{\boldmath$\Theta$\unboldmath}}
\def\muu{\mbox{\boldmath$\mu$\unboldmath}}
\def\Si{\mbox{\boldmath$\Sigma$\unboldmath}}
\def\Gam{\mbox{\boldmath$\Gamma$\unboldmath}}
\def\gamm{\mbox{\boldmath$\gamma$\unboldmath}}
\def\si{\mbox{\boldmath$\sigma$\unboldmath}}
\def\Lam{\mbox{\boldmath$\Lambda$\unboldmath}}
\def\De{\mbox{\boldmath$\Delta$\unboldmath}}
\def\vps{\mbox{\boldmath$\varepsilon$\unboldmath}}
\def\Up{\mbox{\boldmath$\Upsilon$\unboldmath}}
\def\Lap{\mbox{\boldmath$\LM$\unboldmath}}
\newcommand{\ti}[1]{\tilde{#1}}

\def\tr{\mathrm{tr}}
\def\etr{\mathrm{etr}}
\def\etal{{\em et al.\/}\,}
\def\argmax{\mathop{\rm argmax}}
\def\argmin{\mathop{\rm argmin}}
\def\vec{\text{vec}}
\def\cov{\text{cov}}
\def\dg{\text{diag}}

\newtheorem{lemma}{Lemma}
\newtheorem{definition}{Definition}
%\newtheorem{problem}{Problem}
%\newtheorem{proposition}{Proposition}
%\newtheorem{cor}{Corollary}
%------------------------------------------------------------------------------------
\title{Causal Discovery from Incomplete Data: A Deep Learning Approach}
\author{Yuhao Wang$^1$, Vlado Menkovski$^1$, Hao Wang$^2$, Xin Du$^1$, Mykola Pechenizkiy$^1$ \\
$^1$Eindhoven University of Technology, $^2$Massachusetts Institute of Technology\\
\{y.wang9, v.menkovski, x.du, m.pechenizkiy\}@tue.nl, hoguewang@gmail.com
}

%------------------------------------------------------------------------------------
\maketitle
\begin{abstract}
\begin{quote}

As systems are getting more autonomous with the development of artificial intelligence, it is important to discover the causal knowledge from observational sensory inputs. By encoding a series of cause-effect relations between events, causal networks can facilitate the prediction of effects from a given action and analyze their underlying data generation mechanism. 
However, missing data are ubiquitous in practical scenarios. Directly performing existing casual discovery algorithms on partially observed data may lead to the incorrect inference. 
To alleviate this issue, we proposed a deep learning framework, dubbed Imputated Causal Learning (ICL), to perform iterative missing data imputation and causal structure discovery. Through extensive simulations on both synthetic and real data, we show that ICL can outperform state-of-the-art methods under different missing data mechanisms.

\end{quote}
\end{abstract}

%------------------------------------------------------------------------------------
\section{Introduction}

\label{s:intro}
Analyzing causality is a fundamental problem to infer the causal mechanism from observed data. Usually causal relations among variables are described using a Directed Acyclic Graph (DAG), with the nodes representing variables and the edges indicating probabilistic relations among them. 
Learning such causal networks has proven useful in various applications, ranging from smart cities to health care. For example, knowledge of the causal structure is (1) helpful for analyzing relations among different business entities and supporting business decisions \cite{borb}, 
(2) necessary in learning gene regulatory network and analyzing complex disease traits~\cite{wang2017permutation}, (3) important for visualizing causal attentions of self-driving cars, where a highlighted region would causally influence the vehicular steering control \cite{lopez2017discovering}.
In short, the discovered causal networks enable accurate decision making \cite{sulik2017encoding}, robust uncertainty inference \cite{nakamura2007information}, reliable fault diagnose \cite{cai2017bayesian}, and efficient redundancy elimination \cite{xie2017anomaly}.

Previous works on causal discovery mainly focus on the complete-data setting. They either try to learn the Bayesian network structure to estimate Markov properties or use the addictive noise model for causal inference. However, causal discovery under the missing-data setting is still relatively under-explored~\cite{Gain2018a}. In practice, missing data is a common issue. The underlying missingness mechanisms can be categorized into three basic types: Missing At Random (MAR), Missing Completely At Random (MCAR), and Missing Not At Random (MNAR). 
For example, sensors on the road intersection can record the traffic density, and traffic related information will be transmitted to the Road Side Units (RSUs) for traffic management in real time. In MAR, missingness is caused by fully observed variables. For example, when the vehicle density is above a threshold, RSUs will get overloaded and fail to collect traffic data. Missing traffic data depends on the traffic density recorded by the traffic sensor.
MCAR is a special case of MAR, the cause of missingness is purely random and does not depend on the variables of interest, such as the lost of traffic information happens by chance.
In MNAR, missingness depends on either unobserved attributes or the missing attribute itself. For example, the missingness of RSUs depends on the traffic density detected by the sensor. Additionally, the sensor itself also introduces missing values.

Some of the previous approaches handling missing data by directly deleting data entries with missing values, resulting in a complete observation for the problem at hand \cite{carter2006solutions,van2015efficient}. This data processing way may be satisfactory with a small proportion of missing values (e.g., less than about 5\% \cite{graham2009missing}), but could result in a biased model in the presence of larger missing proportions. In theory, MCAR and MAR conditions ensure the recoverability of the underlying distributions from the measured value alone \cite{nakagawa2015missing}, and do not require the prior assumption of how data are missing. Therefore, a feasible solution can be first performing imputation to recover the missing entries, then followed by a causal discovery algorithm for knowledge representation from the recovered data \cite{strobl2018fast}. However, as will be discussed further, directly perform imputation could introduce incorrect causal relations.

In this paper, we focus on causal discovery from observational data (as opposed to intervention experiments). Note that estimating the causal graph as a DAG is an NP-complete problem \cite{chickering1996learning}, and the task becomes even more challenging under the missing data condition. Causal discovery is an unsupervised learning problem and the goal is to discover the data generation process in the form of causal graphs. 
Inspired by \cite{yu2019dag} and motivated by the recent success of Generatvie Adversarial Networks (GAN) \cite{goodfellow2014generative} and Variational Autoencoder (VAE) \cite{kingma2013auto} in learning high-dimensional distributions, in this work, we use GAN and VAE to decompose this problem into two sub-problems, namely, iterative imputation with causal skeleton learning, and identify individual pairs of causal directions.
In general, causal skeleton learning returns a reasonable network structure and offers a global view of how variables are dependent on each other, while causal direction identification provides a more accurate local view between the matched variable pairs. These complimentary local and global view helps approximate the data generating process among all observed variables.

Our contribution is three-fold:
\begin{compactitem}
    \item We propose a deep learning framework, called Imputed Causal Learning (ICL), for iterative missing data imputation and causal structure discovery, producing both imputed data and causal skeletons. 
    \item We leverage the extra asymmetry cause-effect information within dependent pair sets in the causal skeleton $\widetilde{\mathcal{G}}$. 
    The causal directions in $\widetilde{\mathcal{G}}$ then being enumerated in a pair-wise way to uncover the underlying causal graph $\mathcal{G}$. 
    \item Through extensive simulations on both synthetic and publicly-used real data, we show that under MCAR and MAR conditions, our proposed algorithm outperforms state-of-the-art baseline methods.
\end{compactitem}

%------------------------------------------------------------------------------------
\section{Related Work}

\subsubsection*{\textbf{Causal Discovery from Complete Data}}
\textblack{
Methods for identifying causal relations from complete observation data usually fall into two categories: the first one exploits Markov properties of DAGs \cite{chickering2002}, and the second one tries to leverage asymmetries between variable pairs of the Functional Causal Model (FCM) \cite{shimizu2006,mooij2016}.
For methods in the first category, they may not be able to orient the causal direction of $X - Y$, since $X \rightarrow Y$ and $Y \rightarrow X$ are Markov equivalent. However, the causal direction can be further identified using methods in the second category by leveraging the asymmetry between causes and effects.} 
Methods in the first category typically include constraint-based approaches, score-based approaches, and hybrid approaches. They can discover the dependence relations and identify the Markov equivalence class. \textbf{Constraint-based approach} discovers conditional independence between variables of DAGs. Typical algorithms under this category include the PC algorithm, Fast Causal Inference (FCI) \cite{spirtes2000}, and Really Fast Causal Inference (RFCI) \cite{colombo2011learning}. Greedy Equivalence Search (GES) \cite{nandy2018high} is a \textbf{Score-based approach}, it performs structure learning 
with a scoring criteria over the search space of the Markov Equivalence class. The recent breakthrough \cite{zheng2018dags} makes the score-based method amenable with the existing black-box solvers. 
DAG-GNN \cite{yu2019dag} learns the DAG structure using a graph neural network.
Besides, \textbf{hybrid approaches}, such as the the Adaptively Restricted Greedy Equivalence Search (ARGES) \cite{nandy2018high}, Causal Generative Neural Network \cite{goudet2018learning}, which combine ideas of constraint and score-based approach. They restricts the score-based search space with the help of the conditional independence graph for either the computational efficiency or performance accuracy. 
Meanwhile, methods in the second category can be used to identify the causal directions, include linear non-Gaussian acyclic model (LiNGAM) \cite{shimizu2006},
Addictive Noise Model (ANM) \cite{peters2014causal}, Post-nonlinear model (PNL) \cite{zhang2016estimation}. 
%------------------------
\begin{figure*}
\centerline{\includegraphics[width=168mm,scale=0.5]{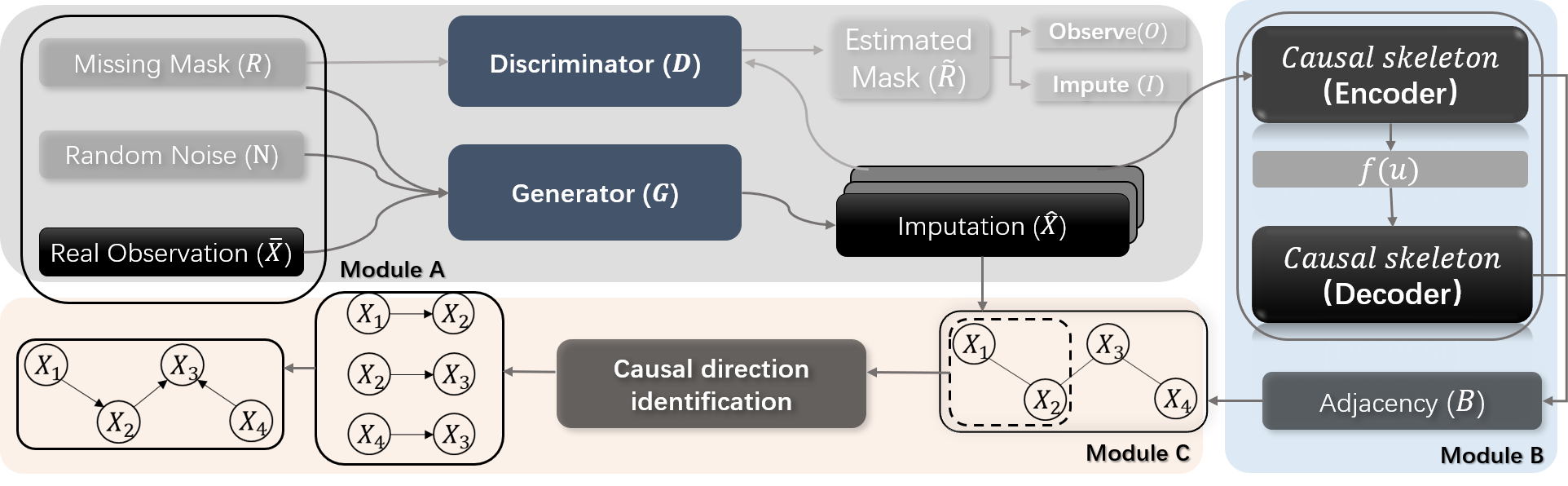}}
\vspace*{-2mm}
\caption{System architecture of our proposed ICL network, including three modules. We train Module A and B in an end-to-end manner for simultaneously imputation and causal skeleton learning, and the results is used as the input of Module C for causal direction identification.}
\vspace*{-5mm}
\label{fig:system_framework}
\end{figure*}
%------------------------

\subsubsection*{\textbf{\textblack{Causal Discovery from Incomplete Data}}}

Works related to causal discovery from incomplete data can be classified into two categories: one category attempts to discover causal structure using only available partial observations and the other aims at imputing all missing entries to recover the whole observation. 
Typical algorithms with partial observations perform (1) list-wise deletion on all entries (rows) with missing values before causal discovery \cite{pmlrv72gain18a}. (2) Test-wise deletion effectively ignores only the variables containing missing values involved in the conditional independence (CI) test \cite{strobl2018fast,tu2019causal}. 
These methods are suitable when the missingness mechanism can not be ignored and the underlying distribution is less likely to be recovered. 
Another category attempts to impute the missing values before performing causal discovery. Previous works use Expectation Maximization (EM) or Gibbs sampling to perform imputation. 
However, these approaches require prior knowledge of the underlying structure and are therefore not practical~\cite{singh1997learning}. On the other hand, imputation strategies for handling missing data is also very important. Works related to this category include the Multivariate Imputation by Chained Equations (MICE) \cite{white2011multiple}, MissForest (MF) \cite{stekhoven2011missforest}, and deep-learning-based approaches, such as using GAN for more powerful imputation \cite{li2018learning,luo2018multivariate,yoon2018gain}. 
In this context, recovering the full distributions from missing data through imputation and performing causal discovery on the recovered data is the most straightforward solution \cite{adel2017learning}.
%------------------------------------------------------------------------------------

\section{Imputed Causal Learning}
\label{sec:icl}

On a high level, our model first takes incomplete observational data $\bar{X}$ as input and then simultaneously performs missing data imputation and structural learning to estimate both the causal skeleton (as an undirected graph) and the recovered data $(\widetilde{\mathcal{G}},\hat{X})$ (Module A and B of \Cref{fig:system_framework}). After that, pair-wise causal direction identification is performed to orient the causal edges and uncover the final underlying causal graph $\mathcal{G}$ (Module C of \Cref{fig:system_framework}). \Cref{fig:system_framework} shows an overview of our framework. The following subsections explain these two steps in detail.

\subsubsection*{\textbf{Notation and Preliminaries}}

A causal graph $\mathcal{G}=(\mathcal{V,E}) $ consists of nodes $\mathcal{V}$ and edges $\mathcal{E}$. 
We denote a random variable set $X$ with $X := (X_{1},X_{2},...,X_{d}), X \in \mathbb{R}^{n \times d}$ to represent $n$ i.i.d. observations into an $n \times d$ data matrix.
Node set $\mathcal{V}$ corresponds to $d$ vertices,
whereby each node $i \in \mathcal{V}$ in $\mathcal{G}$ represents a random variable $X_{i}$ in a causal DAG. 
Within the edge set $\mathcal{E}$, an edge from two adjacent nodes $X_{i}$ to $X_{j}$ exists if and only if $(i,j)\in \mathcal{E}$ and $(j,i)\notin \mathcal{E}$, leading to a cause-effect pair of $X_i \rightarrow X_j$. \textblack{A causal skeleton can be represented as $X_i - X_j$.}
Besides, linear causal relationship in a form of graph $\mathcal{G}$ can be equivalently represented as a linear Structural Equation Model (SEM):
\begin{equation}
\small
    X_j = \sum_{K \in pa_{j}^{\mathcal{G}}}\beta_{ij} X_i + u_j \:\:\: \:(j=1,...,d).
\label{eq:1}
\end{equation}
And the relations between variables in rows are equivalent to $X = B^TX + U$. 
$B \in \mathbb{R}^{d \times d}$ is a strictly upper triangular adjacency matrix with $B_{i,i}=0$ for all $i$, and $B_{i,j} \neq 0$ represent an edge between $X_i$ and $X_j$ in $\mathcal{G}$. $U$ is an $n\times d$ noise matrix with noise vectors $U := (u_1,u_2,...,u_d)$.
Furthermore, a generalized nonlinear SEM model can be formulated as $X=B^T f(X)+U$ \cite{yu2019dag}. $B^T$ can be treated as an autoregression matrix of the DAG. The joint probability distribution $P(X)$ is defined over the graphical model with a finite set of vertices $\mathcal{V}$ on random variables $X$.

\subsection{Causal Skeleton Discovery from Incomplete Data}

\subsubsection*{\textbf{Problem Formulation and Method Overview}}

Under the missing data condition, we assume confounders (unobserved direct common cause of two variables) do not exist in the input data. This means that we can observe all variables but some samples may be missed. We define an incomplete version of $X$ as $\bar{X}:=(\bar{X}_1, \bar{X}_2, ...,\bar{X}_d)$, where $R = (R_1, R_2,...,R_d)$ in Equation (\ref{eq:2}) is the corresponding masks. $ R\in{\{0,1\}}^{d}$ is a binary random variable and used to denote which entries in $\bar{X}$ are missing. Specifically:
\begin{equation}
\small
    \bar{X}_i=\begin{cases}
               X_i, & if\;R_i = 1;\\
               *\;, & otherwise,\\
            \end{cases}
\label{eq:2}
\end{equation}
where $*$ means `missing'. 

In this paper, causal skeleton discovery from incomplete data refers to the problem of inferring $B$ from incomplete observations $\bar{X}$. We do this by iteratively imputing $\bar{X}$ and updating $B$.

\textbf{Imputing $\bar{X}$}: Note that unlike previous causal discovery approaches dealing with missing data by either list-wise or test-wise deletion, we aim to generate full observations and yield an optimistic estimation from $\bar{X}$ by imputation. 
Therefore, with $\bar{X}$ only, we then need to first recover the underlying joint probability distribution $P(X)$ from $\bar{X}$, and representing $P(\bar{X})$ with a structured dependency among variables in $\bar{X}$ with $P(\bar{X})=\prod_{i}(\bar{X_i}|PA_i)$, where $PA_i$ denotes the set of parents of node $i$.
We denote the recovered data by $\hat{X} \in \mathbb{R}^{n \times d}$, and then formulate our task as minimizing the distribution difference of $P(X)$ and $P(\bar{X})$ by imputing all missing values of $\bar{X}$ into $\hat{X}$. 

\textbf{Updating $B$}: In each iteration after imputing $\bar{X}$, we infer (and update) the autoregression parameter $B$ with $\hat{X}=B^T f(\hat{X})+U$ by mapping samples from $\hat{X}$ into a linear-separable hyperspace of $f(\hat{X})$ with a neural network. 

\textbf{Iterative Update}: The imputation (Module A of \Cref{fig:system_framework}) and learning of $B$ (Module B of \Cref{fig:system_framework}) are performed jointly. This is important since the data imputation and learning of $B$ can adjust and improve each other.

%-------------------------------------
\begin{figure}
    \centering
      \begin{subfigure}{0.22\textwidth}
        \includegraphics[width=0.95\textwidth]{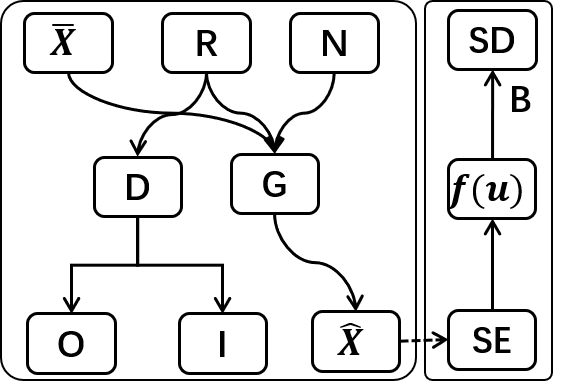}
          \caption{ }
          \label{fig:subfig_a} 
      \end{subfigure}
      \begin{subfigure}{0.22\textwidth}
        \includegraphics[width=0.95\textwidth]{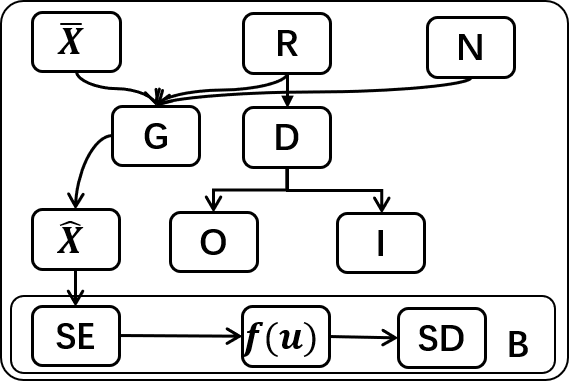}
          \caption{ }
          \label{fig:subfig_b} 
      \end{subfigure}
%\vspace{-3mm}
\vskip -0.3cm
\caption{Incomplete data causal structure discovery (a) Imputation first, then structure discovery; (b) Simultaneous imputation and structure learning;
\label{fig:method_compare}
}
\vspace{-5mm}
\end{figure}
%-------------------------------------

\subsubsection*{\textbf{Proposed Method}}

Built on GAN and VAE, we \textblack{generalize the work on Bayesian structure learning from \cite{yu2019dag} and} propose a deep learning algorithm to simultaneously perform missing data imputation and causal skeleton discovery. Our algorithm consists of four components: a generator ($G$), a discriminator($D$), a structure encoder ($SE$), and a structure decoder ($SD$).
Given incomplete observational data $\bar{X}$, $G$ and $D$ learn to obtain the estimated complete data $\hat{X}$, based on which $SE$ and $SD$ will try to learn the causal structure $B$. These four components are trained jointly using backpropagation (BP).

Note that a naive approach would be to perform imputation first and  \textcolor{black}{then follow by causal discovery} (\Cref{fig:subfig_a}). This is sub-optimal because the estimated causal discovery cannot improve the data imputation process in turn. Empirically we find that its performance is very similar to performing causal discovery after directly deleting all data entries with missing values, meaning that imputation does not introduce any additional value into the causal discovery process. We address this issue by alternating between imputation and causal discovery, which is made possible through the use of differentiable neural networks (\Cref{fig:subfig_b}). 
\textblack{Such an iterative process can do better in terms of performing multiple imputation passes to take into account the variability while preserving the underlying statistical causal relationship between variables.}

Concretely, in each iteration of our algorithm, $G$ and $D$ take the incomplete data as input and impute the missing values to form $\hat{X}$. The causal structure $B$ is involved as parameters of both $SE$ and $SD$. We encode $\hat{X}$ into a latent code $f(U)$ through $SE$, and decode $f(U)$ into $\Tilde{X}$ with $SD$. 
The above procedure can be seen as two neural network modules, GAN and VAE, jointly learning together. The former recovers missing data while the later discovers the causal structure.

\subsubsection*{\textblack{Missing Data Imputation}}
Similar to~\cite{luo2018multivariate,yoon2018gain}, we use $G$ and $D$ together to approach the underlying data distribution of $P(X)$ for imputation. 
Since GAN can not take $NaN$ values as the input, to initialize the imputation process, we use a d-dimensional noise variable $N=(N_1, N_2, ..., N_d)$ sampled from the standard normal distribution $N\sim \mathcal{N}(\0 , \I)$. And we replace the $NaN$ entries in $\bar{X}$ with $\bar{X}=R \odot \bar{X} +  (1-R)\odot N$, where $\odot$ represent element-wise multiplication. 
$\bar{X}$ will be served as the input of GAN to generate $\hat{X}$.
With $\hat{X}$ as the input of the structure learning neural network of $SE$ and $SD$ to discover the autoregression parameter $B^T$ through each iteration (details of the structure learning method will we covered in the next subsection). Specifically, 
the generator is responsible for observing the real data and imputing missing components conditioned on what is actually observed according to $P(\hat{X}|\bar{X})$. The generator takes $\bar{X}$, $R$, and $N$ as input:
\begin{equation}
\small
    \tilde{X} = G(R, \bar{X}, (1-R)\odot N).
\label{eq:3}
\end{equation}
Therefore, the recovered data $\hat{X}$ can be obtained by replacing data on missing entries in $\bar{X}$ with the generated corresponding values from $\tilde{X}$ as
\begin{equation}
\small
    \hat{X}=R \odot\bar{X} +   (1-R)\odot \tilde{X}.
\label{eq:4}
\end{equation}

Besides, the discriminator $D$ is introduced as an adversary training module accompanying the generator $G$. Due to the incomplete observations, the initialized data $\bar{X}$ inherently contains both real and fake values, which makes $D$ from standard GAN not feasible for our task. In this context, instead of counting real/fake from $\tilde{X}$, the mapping of $D(\cdot)$ attempts to determine whether the components are actually observed or not. Specifically, we set $\hat{X}$ as the input to $D$, while $G$ is trying to fool $D$ in an adversarial way.
In summary, $G$ and $D$ together learn a desired joint distribution of $P(\hat{X})$ and then perform imputation given $\bar{X}$. Note that the difference between ICL and previous GAN-based imputation methods is that our imputation is also related to the recovered causal skeleton.

\subsubsection*{\textblack{Causal Skeleton Learning}}

Then we perform structure discovery to find the underlying causal mechanism from the variable set $\mathcal{V}$ in $\hat{X}$.
Using the structure discovery method in \cite{yu2019dag}, with the scoring function $\mathcal{S_D}$, this concatenate task then turns into a continuous optimization problem of finding a $\widetilde{\mathcal{G}}$ that satisfies: 
\begin{equation}
\begin{split}
\small
    & \widetilde{\mathcal{G}} = g(argmin_{\mathcal{G}\in \mathbb{R}^{d \times d}}\;{\mathcal{S_D}}(\mathcal{G})),\\
    & s.t. \:\:h(\mathcal{G})=tr[(I+\alpha B\circ B)^d]-d=0,
\end{split}
\label{eq:5}
\end{equation}
where $g(\cdot)$ is a function to remove directions in $\mathcal{G}$, leading to predicted causal skeleton $\widetilde{\mathcal{G}}$. The adjacency matrix space $\mathbb{R}^{d \times d}$  represents the set of all DAGs. $h: \mathbb{R}^{d \times d} \rightarrow \mathbb{R}$ is the smooth function over real matrices, and $h({\mathcal{G}})=0$ ensures that $\mathcal{G}$ is acyclic. 
$\alpha$ is a hyperparameter. Following \cite{yu2019dag}, $SE$ takes $\hat{X}$ as the input of a multilayer perceptron (MLP). The output, denoted as $MLP(\hat{X}, \textbf{W}_1)$, is then multiplied by $(I-B^T)$ and transformed into $f(U)$ in Equation \eqref{eq:6}, where $U$ is the noise vector mentioned at the start of Section~\ref{sec:icl}. The decoder $SD$ in Equation \eqref{eq:7} performs an inverse operation $(I-B^T)^{-1}$ on the encoded $f(U)$ to recover $\Tilde{X}$, where $B$ is a parameter of $SE$ and $SD$ to incorporate the causal structure during the learning process.
$I$ denotes the identity matrix. $\textbf{W}_1$ and $\textbf{W}_2$
are parameters in corresponding layers. 
\begin{gather}
\small
    f(U)=(I-B^T)MLP(\hat{X}, \textbf{W}_1); \label{eq:6}\\
    \Tilde{X} = MLP((I-B^T)^{-1}(f(U), \textbf{W}_2)), \label{eq:7} 
\end{gather}
The parameter $B$ plays an important role during the learning phase, $B_{i,j} \neq 0$ stands for the dependence relationship between $X_i$ and $X_j$ in $\widetilde{\mathcal{G}}$. 

By extracting $B^T$ from the learning process described in Equations \eqref{eq:6} and \eqref{eq:7}, we can have the knowledge of the marginal or conditional distribution of a random variable in $\mathcal{V}$. This is also how we discover a causal skeleton from $\hat{X}$.

\subsubsection*{\textblack{Joint Training}}

The overall procedure can be seen as simultaneously recovering all missing entries in $\bar{X}$ by $G$ and $D$, and optimizing the structure learning performance of {$P(\widetilde{\mathcal{G}}|\hat{X}, R)$} by $SE$ and $SD$. 

The loss function is formed into two parts as the imputation loss and structure learning loss. Since the missing values in real-scene are not known, it would not make sense to use their reconstruction error as a stopping criterion in the imputation loss part. The training objective can be formulated as a $minimax$ problem of $\underset{G}{min}\:\mbox{ } \underset{D}{max}\: L_i(D,G)$ while measuring the degree of imputation fitness, as it is usually done when using the standard GANs. In our work, we optimize the data generation performance of a GAN with the loss function as follows
\begin{equation}
\small
\begin{split}
    & L_{i}(D, G)=\mathbb{E}_{\bar{X},R,N}[R^T\log D(G(R, \bar{X}, (1-R)\odot N)) \\  &+(1-R^T)\log(1-D(G(R, \bar{X}, (1-R)\odot N)))].
\end{split}
\label{eq:8}
\end{equation}
The generator $G$ generates samples conditioned on the partial observation of $\bar{X}$, the missingness indicator $R$, and the noise $N$. We train $G$ to generate $\hat{X}$ and minimize the prediction probability of $R$, while we train $D$ to maximize the prediction accuracy of $R$. Then we follow the evidence lower bound (ELBO) from \cite{yu2019dag}, given below, for causal skeleton learning.

\begin{equation*}
\small
    L_{e} =  -\mathbb{E}_{q{(U|\hat{X})}}[\log p{(\hat{X}|U)}] + D_{KL}(q{(U|\hat{X})|| p{(U))}}
\label{eq:9}
\end{equation*}
We denote $\Phi$ and $\Theta$ as parameter sets in GANs and VAEs separately. The overall learning problem can be formed as:
\begin{equation}
\small
\begin{split}
     &\underset{\Phi} min\:f(\Phi) =  L_{i}(G, D); \\
     &\underset{B,\Theta}{min}\:f(B, \Theta) = -L_{e},\: s.t.\:\:h(B) = 0.
\end{split}
\label{eq:10}
\end{equation}
The stopping criteria is either the error is sufficiently small or the number of iterations is large enough.
With the best fitting $B$ in Equation \eqref{eq:10}, the causal skeleton $\widetilde{\mathcal{G}}$ is generated by keeping edges in $\mathcal{E}$ but remove their directions. The pseudo code is summarized in Algorithm \ref{alg1}.
%=================================================

\begin{algorithm}
\small
\caption{Causal Skeleton Discovery}
\label{alg1}
  \Initialize {$R \in \{0,1\}^{n\times d}$, $\bar{X}\in\mathbb{R}^{n\times d}$, $\widetilde{\mathcal{G}}\in \mathbb{R}^{d\times d}$, \\$N = P_n\sim \mathcal{N}(\mu ,\sigma^2)$, minibatch $J$.}
  \Input{\text{Observational incomplete data} $\bar{X}$.}
  \Output{Causal skeleton and imputed data $(\widetilde{\mathcal{G}},\hat{X})$.}
  \While{Loss has not converged}{
    \For{$j = 1:J$}{ 
        \textbf{Step 1: \textit{Missing data imputation:}} \\
         Missing entries: $\tilde{X} = G(R, \bar{X}, (1-R)\odot N).$ \\
        Imputation: $ \hat{X}=R \odot\bar{X} +  (1-R)\odot\tilde{X}.$ \\
        \textbf{Step 2:\textit{Structure discovery: }}\\
        $SE$: $f(U)=(I-B^T)MLP(\hat{X}, \textbf{W}).$\\
        $SD$: $\Tilde{X} = MLP((I-B^T)^{-1}(f(U), \textbf{W})).$\\
        \textbf{Step 3: \textit{Extract $\mathcal{G}$ from $B$: }}\\
        Let $\widetilde{\mathcal{G}}=(\mathcal{V,E})$ with $\mathcal{E}=\{(i,j):B_{i,j}\neq0\}.$\\
        \textbf{Step 4}: \textit{Update parameters $\Phi$ of G and D in GAN using SGD according to Equation \eqref{eq:8}}.\\
        \textbf{Step 5}: \textit{Update parameters $\Theta$ of SE and SD in VAE and $B$ using SGD according to Equation \eqref{eq:10}.} } 
    }
\end{algorithm}
%========================================

\subsection{Causal Direction Identification}

The above procedure can identify the conditional probability, but may not truly represent the underlying causal mechanism. For example, given two variables $X_i$ and $X_j$, their joint distribution $P(X_i, X_j)$ can be decomposed equally as either $P(X_j|X_i)P(X_i)\:(X_i\rightarrow X_j)$ or $P(X_i|X_j)P(X_j)\:(X_j \rightarrow X_i)$. 
These two decompositions relate to different causal mechanisms.
With the additive noise model \cite{mooij2016} $\hat{X_j}=f(\hat{X_i})+U\;, U\indep \hat{X_i}$, however, we can represent asymmetries between $X_i\rightarrow X_j$ and $X_j\rightarrow X_i$, leading to a unique causal direction from purely observational data. 
In detail, let the joint distribution of $P(\hat{X_i},\hat{X_j})$ with ground truth be $\{(\hat{X_i} \rightarrow \hat{X_j}),\;(i,j)\in d\}$. Then the effect of $X_j$ conditioned on the cause $X_i$ can be represented by:
\begin{equation}
\small
\begin{split}
    &P(\hat{X}_j=x_{j}^m|\hat{X}_i=x_{i}^m) =\;\frac{P(\hat{X}_j=x_{j}^m, \hat{X}_i=x_{i}^m)}{P(X_i=x_{i}^m)} \\
    &\underset{X_j\not\!\perp\!\!\!\perp U}{\overset{\mathrm{X_i \indep U}}{=\joinrel=}} \frac{P( U=x_{j}^m-f(x_{i}^m)) P(X_i=x_{i}^m)}{\mathbb{P}(X_i=x_{i}^m)} \\
    & \:= \:P(U =x_{j}^m-f(x_{i}^m)) \\
    & \:= \:P(U =\epsilon) ,\: \:(X_i \rightarrow X_j, \:(i,j)\in d, \:m \in n),
\end{split}
\vspace{-5mm}
\label{eq:11}
\end{equation}
where the second equality assumes $X_j\not\!\perp\!\!\!\perp N$ and $X_i \indep N$. Note that due to the asymmetry, Equation \eqref{eq:11} does not hold in the reverse direction $X_j \rightarrow X_i$. This property makes it possible to determine the causal direction from observational data under proper conditions.

Therefore, given $(\widetilde{\mathcal{G}},\hat{X})$ from the above section, our goal is to utilize such pair-wise asymmetry and orient the edges of $\widetilde{\mathcal{G}}$, consequently uncovering the final causal DAG $\mathcal{G}$. 
This can be achieved by calculating the maximum evidences of the marginal log-likelihood over two models $M(X_i, X_j)$ and $M(X_j, X_i)$. The model that shows the larger evidence is selected. In this work, we use the Cascade Additive Noise Model (CANM) proposed by \cite{ijcai2019-223}. 
Specifically, to enumerate causal direction from variables pairs in $\widetilde{\mathcal{G}}$, we use variable pairs $\hat{X}(x_{i}^m, x_{j}^m)$ from $\hat{X}$ as input, then the log-marginal likelihood on variable $X_i$ and $X_j$ is computed with:
\begin{equation*}
\small
\begin{aligned}
    & \log p_{\theta}{(X_i, X_j)} = 
    \log\prod_{m=1}^{n}\int p_{\theta}(\hat{x}_i^m,\hat{x}_j^m,z)dz \\
    & := \sum_{m=1}^{n}\mathcal{L}(\theta, \phi; \hat{x}_i^m,\hat{x}_j^m) +
    KL(q_{\phi}(z|\hat{x}_i^m, \hat{x}_j^m)\parallel p_{\theta}(z|\hat{x}_i^m, \hat{x}_j^m)) \\ 
    & \geq \sum_{m=1}^{n}\mathcal{L}(\theta, \phi; \hat{x}_i^m,\hat{x}_j^m).
\end{aligned}
\label{eq:12}
\end{equation*}
$\theta$ and $\phi$ are the parameters of the CANM model, which encode $\hat{x}_i^m$ and $\hat{x}_j^m$ into a latent code $z$.
The evidence score ${S}_{x_i \rightarrow x_j}$ of the log marginal likelihood with  $ \sum_{m=1}^{n}\mathcal{L}(\theta, \phi; \hat{x}_i^m,\hat{x}_j^m)$ can be calculated in the following way in both directions.
\begin{equation*}
    \sum_{m=1}^{n} E_{z \sim {q_{\phi}(z|x_i,x_j)}} [-\log q_{\phi}(z|x_i, x_j) + \log p_{\theta}(x_i,x_j,z)].
\label{eq:13}
\end{equation*}
\noindent And the causal direction can be identified by:
\begin{equation}
\text{dir\::=}
\begin{cases}
 \hat{X_i} \rightarrow \hat{X_j} , &if\:\hat{S}_{x_i \rightarrow x_j} >  \hat{S}_{x_j \rightarrow x_i} \\
 \hat{X_j} \rightarrow \hat{X_i} , &if\:\hat{S}_{x_i \rightarrow x_j} <  \hat{S}_{x_j \rightarrow x_i} \\
 Not\;determined. &others
\end{cases}
\label{eq:14}
\end{equation}
Given the bivariate identifiable condition in Equation \eqref{eq:11}, causal discovery from more than two variables can be achieved if each of the causal pairs follows the ANM class \cite{peters2011identifiability}. 
To uncover the underlying causal graph $\mathcal{G}$, we then independently orient each pair-wise edge using the bivariate identification method in Equation (11).
Besides, note that a combination of causal structure learning and bi-variate direction identification requires a final verification to ensure that the DAG is acyclic. In the final stage, by checking if cycles $\mathcal{G_C}$ in $\mathcal{G}$ exist, we enumerate the related edges with the calculated score $(\mathcal{E}_{ij}, S_{x_i,x_j})$, then simply remove the edge which holds the lowest score. We will consider more sophisticated algorithms in future work.

\begin{table*}[]
\vskip -0.4cm
%\scriptsize
\footnotesize
\centering
\caption{\small Performance comparison (mean and standard deviation) using Structural Hamming Distance, lower is better.}
\vskip -0.2cm
\begin{tabular}{c|l|ccc|ccc}
\hline
\multicolumn{2}{c|}{\multirow{2}{*}{}} & \multicolumn{3}{c|}{30 Var MCAR (Nonlinear 1) (Ideal SHD=7)}                                                & \multicolumn{3}{c}{50 Var MAR (Nonlinear 2) (Ideal SHD=17)}                                                \\ \cline{3-8} 
\multicolumn{2}{c|}{}                  & \multicolumn{1}{c}{10\%} & \multicolumn{1}{c}{30\%} & \multicolumn{1}{c|}{50\%} & \multicolumn{1}{c}{10\%} & \multicolumn{1}{c}{30\%} & \multicolumn{1}{c}{50\%} \\ \hline
\multirow{3}{*}{GES}     & LD-GES      & $106.0 \pm 14.3$         & $109.1 \pm 16.9$         & $145.4 \pm 13.4$         & $227.2 \pm 22.5$         & $224.1 \pm 28.6$         & $225.6 \pm 28.4$        \\ %\cline{2-8} 
                         & GAN-GES     & $107.8 \pm 12.2$         & $106.9 \pm 14.8$         & $133.1 \pm 15.9$         & $228.5 \pm 21.3$         & $224.2 \pm 25.6$         & $225.6 \pm 27.8$        \\ %\cline{2-8} 
                         & MF-GES      & $109.3 \pm 13.8$         & $108.1 \pm 14.8$         & $136.9 \pm 16.1$         & $230.6 \pm 21.6$         & $224.1 \pm 28.5$         & $223.9 \pm 26.9$        \\    
                         & MC-GES      & $109.3 \pm 13.8$         & $109.1 \pm 15.2$         & $132.3 \pm 16.2$         & $230.6 \pm 21.6$         & $225.4 \pm 28.0$         & $225.4 \pm 27.2$        \\ \hline
\multirow{4}{*}{RFCI}    & LD-RFCI     & $22.2 \pm 5.2$           & $26.4 \pm 8.3$           & $43.3 \pm 7.4$           & $44.1 \pm 8.3$           & $49.7 \pm 8.8$           & $68.2 \pm 10.1$         \\ %\cline{2-8} 
                         & GAN-RFCI    & $38.6 \pm 5.1$           & $39.9 \pm 8.3$           & $42.0 \pm 7.3$           & $52.3 \pm 8.3$           & $66.6 \pm 8.7$           & $69.2 \pm 10.1$         \\ %\cline{2-8} 
                         & MF-RFCI     & $38.9 \pm 5.0$           & $39.9 \pm 8.3$           & $44.6 \pm 7.0$           & $51.0 \pm 8.4$           & $66.7 \pm 8.8$           & $68.8 \pm 9.7$          \\ %\cline{2-8} 
                         & MC-RFCI     & $38.8 \pm 4.8$           & $39.8 \pm 8.3$           & $42.7 \pm 7.1$           & $51.7 \pm 8.2$           & $66.5 \pm 9.1$           & $69.0 \pm 10.1$         \\ \hline
\multirow{4}{*}{LiNGAM}  & LD-LiNGAM   & $22.0 \pm 8.4$           & $25.3 \pm 10.3$          & $32.6 \pm 10.4$          & $41.3 \pm 15.2$          & $50.4 \pm 17.6$          & $53.9 \pm 7.1$          \\ %\cline{2-8} 
                         & GAN-LiNGAM  & $20.9 \pm 8.4$           & $23.1 \pm 10.3$          & $37.0 \pm 10.4$          & $43.0 \pm 15.2$          & $53.2 \pm 17.6$          & $47.6 \pm 7.1$          \\ %\cline{2-8} 
                         & MF-LiNGAM   & $23.1 \pm 7.8$           & $23.5 \pm 8.3$           & $37.6 \pm 11.2$          & $52.0 \pm 16.9$          & $48.2 \pm 18.1$          & $52.4 \pm 13.6$         \\ %\cline{2-8} 
                         & MC-LiNGAM   & $21.5 \pm 8.9$           & $29.1 \pm 12.3$          & $37.3 \pm 12.0$          & $43.6 \pm 13.1$          & $51.9 \pm 14.0$          & $52.6 \pm 11.2$         \\ \hline
\multirow{4}{*}{PC}      & LD-PC       & $26.2 \pm 6.2$           & $27.9 \pm 7.6$           & $35.0 \pm 6.4$           & $36.0 \pm 7.7$           & $38.5 \pm 10.4$          & $45.2 \pm 8.1$          \\ %\cline{2-8} 
                         & GAN-PC      & $26.0 \pm 6.2$           & $26.1 \pm 7.6$           & $32.3 \pm 6.4$           & $34.2 \pm 7.7$           & $38.6 \pm 10.4$          & $41.6 \pm 7.4$          \\ %\cline{2-8} 
                         & MF-PC       & $26.4 \pm 5.8$           & $26.2 \pm 7.9$           & $33.3 \pm 6.8$           & $35.0 \pm 8.0$           & $35.3 \pm 10.1$          & $41.9 \pm 7.0$          \\ %\cline{2-8} 
                         & MC-PC       & $27.9 \pm 5.9$           & $26.8 \pm 8.2$           & $33.3 \pm 7.2$           & $34.7 \pm 8.0$           & $37.8 \pm 10.9$          & $42.2 \pm 7.5$          \\ \hline
\multirow{4}{*}{MMPC}    & LD-MMPC     & $22.6 \pm 7.3$           & $23.2 \pm 7.5$           & $30.7 \pm 9.7$           & $45.2 \pm 11.4$          & $44.5 \pm 11.1$          & $44.0 \pm 7.0$          \\ %\cline{2-8} 
                         & GAN-MMPC    & $22.0 \pm 7.5$           & $23.8 \pm 7.2$           & $27.0 \pm 9.9$           & $46.0 \pm 11.1$          & $48.5 \pm 10.5$          & $44.5 \pm 6.5$          \\ %\cline{2-8} 
                         & MF-MMPC     & $22.8 \pm 7.3$           & $25.0 \pm 7.2$           & $29.1 \pm 9.6$           & $46.3 \pm 11.2$          & $48.7 \pm 11.2$          & $44.5 \pm 6.9$          \\ %\cline{2-8} 
                         & MC-MMPC     & $22.4 \pm 7.3$           & $25.8 \pm 7.2$           & $29.4 \pm 9.5$           & $46.3 \pm 11.1$          & $48.6 \pm 11.2$          & $44.4 \pm 7.1$          \\ \hline
\multirow{3}{*}{DAG}     & LD-DAG      & $12.2 \pm 6.2$           & $13.6 \pm 9.2$           & $20.0 \pm 10.4$           & $30.2 \pm 5.9$           & $32.5 \pm 4.5$           & $37.9 \pm 7.1$          \\ %\cline{2-8} 
                         & GAN-DAG     & $11.0 \pm 7.7$           & $10.3 \pm 6.8$           & $14.4 \pm 8.7$           & $23.4 \pm 5.5$           & $27.7 \pm 3.9$           & $30.5 \pm 4.2$          \\ %\cline{2-8} 
                         & ICL (Ours)        & \textbf{9.8 $\pm$ 3.9}           & \textbf{7.4 $\pm$ 3.8}           & \textbf{8.4 $\pm$ 4.9}           & \textbf{19.0 $\pm$ 4.2}           & \textbf{25.5 $\pm$ 3.8}           & \textbf{27.3 $\pm$ 5.5}           \\ \hline
\end{tabular}
\vspace{-5mm}
\label{tab:Table1}
\end{table*}

%------------------------------------------------------------------------------------
\section{Experiment Results}

In this section, we will demonstrate how ICL performs on two synthetic datasets and one real-world dataset compared to state-of-the-art baselines.

\subsection{\textbf{Baseline Algorithms}}

Algorithms for data imputation include list-wise deletion (LD), multivariate imputation by chained equations (MICE) \cite{white2011multiple}, MissForest (MF) \cite{stekhoven2011missforest}, and GAN from as shown in \Cref{fig:subfig_a}. Algorithms for the causal structure discovery include \textbf{constraint-based approaches} such as PC \cite{spirtes2000}, linear non-Gaussian acyclic model (LiNGAM) \cite{shimizu2006}, really fast causal inference (RFCI) \cite{colombo2011learning}, \textbf{score-based approaches} such as greedy equivalence search (GES) \cite{chickering2002}, \textbf{hybrid approaches} such as max-min parents-children-addictive noise model (MMPC-ANM) \cite{cai2018}, and a deep-learning approach based on DAG-GNN \cite{yu2019dag}. For DAG-GNN we consider two variants: GAN-DAG first performs imputation first and then use the imputation results for structure discovery; LD-DAG first delete all entries with missing values and then perform causal discovery.
Each baseline consists of one data imputation algorithm and one causal discovery algorithm. Therefore we have the following combinations: LD-PC, LD-LiNGAM, LD-RFCI, LD-MMPC, LD-GES; MF-PC, MF-LiNGAM, MF-RFCI, MF-MMPC, MF-GES; MC-PC, MC-LiNGAM, MC-RFCI, MC-MMPC, MC-GES; GAN-PC, GAN-LiNGAM, GAN-RFCI, GAN-MMPC, GAN-GES.
All the baseline algorithms above are implemented using R-packages such as \textbf{bnlearn} \cite{scutari2009}, \textbf{CompareCausalNetworks} \cite{heinze2017}, \textbf{pcalg}~\cite{kalisch2012}, and \textbf{SELF}~\cite{cai2018}. We use \textbf{rpy2}~\cite{gautier2012rpy2} to make the above R-packages accessible from Python and ensure that all algorithms can be compared in the same environment. 
Following~\cite{tu2019causal,strobl2018fast}, we use Structural Hamming Distance (SHD) as the evaluation metric.

\subsection{Quantitative Results}

In this subsection, we first provide on synthetic and real-world datasets in terms of both the causal graphs and the missing machanisms. We then compare ICL with the baselines above on these datasets.

\subsubsection*{\textbf{Synthetic Data Generation}}

The synthetic ground truth graph $\mathcal{G}$ with $d$ nodes is generated randomly using the Erd\H{o}s R\'enyi (ER) model with an expected neighbor size $s=2$. The edge weights of $\mathcal{G}$ are uniformly drawn from $B \sim U(-2,-0.5] \cup U[0.5, 2)$ to ensure that they are non-zero. 
Once $\mathcal{G}$ is generated, the observational i.i.d. data  $\bar{X} \in \mathbb{R}^{n \times d}$ is generated with a sample size $n=\{500,1000\}$ and a variable size $d \in \{30, 50\}$. 
For linear cases, the i.i.d. data is generated by sampling the model $X = B^TX + N$, where $B$ is a strictly upper triangular matrix; similarly for nonlinear cases, the sampled model is described by $X = f(B^TX)+N$. Here the noise $N$ follows either the Exponential or the Gumbel distribution. 
In our work, two different mechanisms are considered in nonlinear cases:
\begin{gather*}
\begin{split}
    &1: \;x = 2sin(B^T(x+0.5\cdot1)) + B^T(x+0.5\cdot1) +u,\\
    &2: \;x = \sqrt{x(B^T(x^2+0.5\cdot1))} +u.
\end{split}
\end{gather*}
In order to achieve a more general comparison in our experiments, the missing data proportions over all the synthetic data are set to be 10\%, 30\%, and 50\%.

\subsubsection*{Missingness Mechanisms}

In this paper we generate synthetic incomplete data using one of the two missingness mechanisms, namely MCAR and MAR, leaving MNAR as future work. For MCAR, the missingness mask $R\in \mathbb{R}^{n \times d}$ is formed by selecting the missing entries from the observational data corresponding to $t_i<\tau \:(t_i\in T)$ with the same probability. Here $T \in  \mathbb{R}^{n \times d} $ is a uniformly distributed random matrix which has the same dimensions as the observational data matrix. A threshold $\tau$ is used as the missingness selection criterion. 
For MAR, the missingness mask $R\in \mathbb{R}^{n \times d}$ is generated based on both the randomly generated graph $\mathcal{G}$ (more details later) and $T$. Specifically, we first randomly sample parent-child pairs, denoted as $S_p=\{(i, j)\}$, from $\mathcal{G}$. $R_{kj}$ is then set to $0$ if there exists an $i$ such that $(i,j)\in S_p$ and $T_{ki} <\tau$. This is to simulate the setting where the missingness of the child node is determined by the (randomly generated) values of its parent nodes.

\subsubsection*{\textbf{Quantitative Experiment Results}}

Table \ref{tab:Table1} reports SHD of our proposed ICL and other baeslines. The results are averaged over twenty random repetitions, with the missing proportion $m \in \{10\%, 30\%, 50\%\}$ and under both MCAR and MAR conditions. We cover two nonlinear mechanisms as mentioned above. In our experiments, the linear results is consistent with the nonlinear results, and are not included due to space constraints. As shown in Table \ref{tab:Table1}, ICL shows superior performance compared with all other baselines. 'Ideal SHD' refers to ICL's performance using \emph{complete} data (no missing values). Recall that GAN-DAG performs data imputaion first and then follow by data causal discovery without the iterative process. As shown in Table \ref{tab:Table1}, GAN-DAG's performance is worse than ICL since its causal module cannot improve the data imputation process in turn.
Interestingly, comparing LD-DAG and ICL, we can see that ignoring entries with missing values may have a negative effect on the performance of causal discovery. Furthermore, GES-based algorithms achieve the worst performance even with only 10\% missing values. 
We can also see that \textblack{MMPC-based algorithms are suitable for nonlinear data, while LiNGAM-based algorithms are suitable for linear data}. 
As expected, directly removing the missing entries leads to worse performance, since it not only reduces the sample size (and consequently throwing away useful information in the observational data), but also introduced a form of selection bias, leading to incorrect causal discovery~\cite{pmlrv72gain18a}.
Furthermore, it is also worth mentioning that by performing missing data imputation and causal discovery separately (like GAN-DAG), the results could be even worse than deletion-based methods. As we discussed, imputation could be helpful for recovering the joint distribution of $P(X)$, but sub-optimal when we want to perform a further step of the distribution decomposition
to discover the underlying causal graph.
In contrast, our ICL model does not have the issues above and can therefore achieve better performance.

\subsubsection{Case Study on AutoMPG}

\begin{figure}
\vskip -0.1cm
    \centering
      \begin{subfigure}{0.22\textwidth}
        \includegraphics[width=0.93\textwidth]{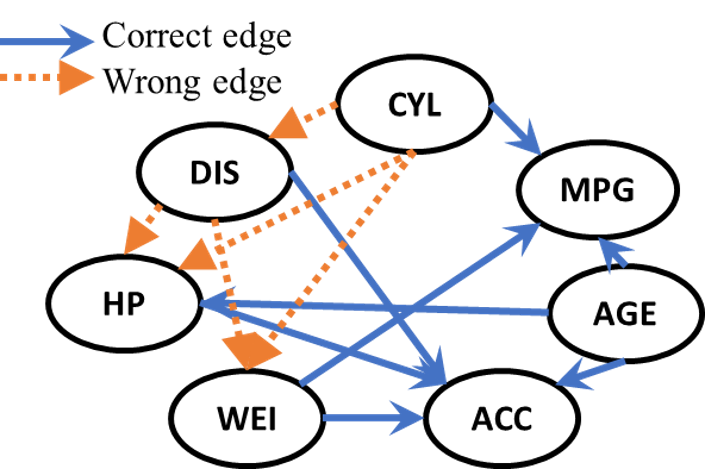}
          \caption{ }
          \label{fig:autompg_a} 
      \end{subfigure}
      \begin{subfigure}{0.22\textwidth}
        \includegraphics[width=0.95\textwidth]{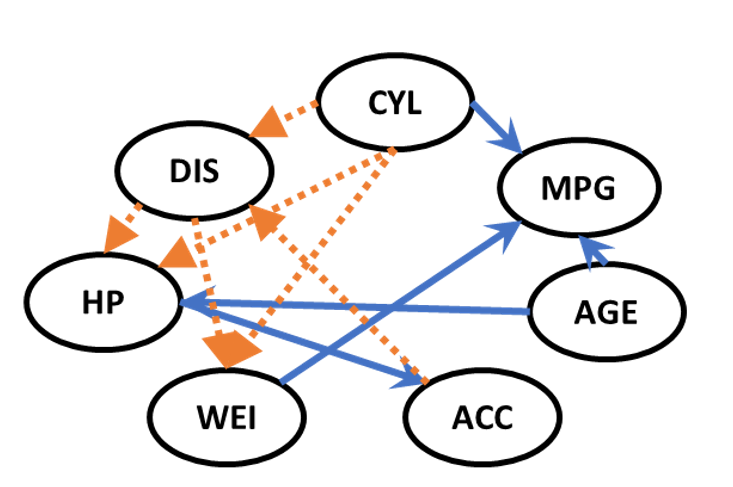}
          \caption{ }
          \label{fig:autompg_b} 
      \end{subfigure}
\vspace{-3mm}
\caption{AutoMPG results. (a) Our ICL algorithm (SHD=9). (b) GAN-DAG (SHD=11).
\label{fig:autoMPG}
}
%\vspace{-5mm}
\vskip -0.6cm
\end{figure}

As a case study we also show ICL's results on a real-world dataset, AutoMPG~\cite{lichman2013uci}, which is a city-cycle fuel consumption dataset with 398 instances. We discard the attributes of the car-name and the origin, and use the left 7 attributes: miles per gallon consumption (MPG), the release date of vehicles (AGE), vehicle weight (WEI), engine displacement (DIS), cylinder number (CYL), horsepower (HP), and vehicle's acceleration capability (ACC).
We simulate $10\%$ missing data under MAR and compare the performance of ICL and GAN-DAG (best baseline). 
Their learned causal networks are shown in \Cref{fig:autoMPG}, where the SHD for ICL and GAN-DAG is 9 and 11, respectively.

%-------------------------------------
%-------------------------------------

\section{Conclusion}

In this work, we addressed the problem of incomplete data causal discovery, and we proposed a deep learning model of ICL to handle this issue. Specifically, our ICL model contains a global view of iterative missing data imputation and causal skeleton discovery, and a local view of enumerating causal directions to uncover the underlying causal $\mathcal{G}$. In the end, we evaluated the effectiveness of our method on both synthetic and real data. As future work, we will generalize our method under more complex conditions such as the existence of confounders.

\bibliographystyle{aaai}
\bibliography{causal_discovery.bib}
\end{document}